%% file: main.tex
\documentclass[sigconf,letterpaper,9pt]{acmart}
\acmConference[FAT*'19]{ACM Conference on Fairness, Accountability, and
Transparency}{January 2019}{Atlanta, Georgia USA}
\acmYear{2019}
\copyrightyear{2019}

\usepackage{booktabs} 

\setcopyright{rightsretained}
\usepackage{tikz}
\usetikzlibrary{bayesnet}
\usepackage{subcaption}
\usepackage{ dsfont }
\usepackage{ mathtools }
\usepackage{ amsmath }
\usepackage{algorithm,algorithmic}
\usepackage{wrapfig}
\usepackage{xcolor}

\DeclareMathOperator*{\argmax}{argmax}

\acmArticle{4}
\acmPrice{15.00}

\begin{document}
\title[Fairness Through Causal Awareness]{Fairness through Causal Awareness: Learning Causal Latent-Variable Models for Biased Data}

\author{David Madras, Elliot Creager, Toniann Pitassi, Richard Zemel}
\affiliation{%
  \institution{University of Toronto, Vector Institute}
}
\email{{madras,creager,toni,zemel}@cs.toronto.edu}

\renewcommand{\shortauthors}{Madras et al.}

\begin{abstract}
How do we learn from biased data?
Historical datasets often reflect historical prejudices; sensitive or protected attributes may affect the observed treatments and outcomes.
Classification algorithms tasked with predicting outcomes accurately from these datasets tend to replicate these biases.
We advocate a causal modeling approach to learning from biased data, exploring the relationship between fair classification and intervention.
We propose a causal model in which the sensitive attribute confounds both the treatment and the outcome.
Building on prior work in deep learning and generative modeling, we describe how to learn the parameters of this causal model from observational data alone, even in the presence of unobserved confounders.
We show experimentally that fairness-aware causal modeling provides better estimates of the causal effects between the sensitive attribute, the treatment, and the outcome.
We further present evidence that estimating these causal effects can help learn policies that are both more accurate and fair, when presented with a historically biased dataset.
\end{abstract}

%
%


\acmYear{2019}\copyrightyear{2019}
\setcopyright{acmcopyright}
\acmConference[FAT* '19]{FAT* '19: Conference on Fairness, Accountability, and Transparency}{January 29--31, 2019}{Atlanta, GA, USA}
\acmBooktitle{FAT* '19: Conference on Fairness, Accountability, and Transparency, January 29--31, 2019, Atlanta, GA, USA}
\acmPrice{15.00}
\acmDOI{10.1145/3287560.3287564}
\acmISBN{978-1-4503-6125-5/19/01}


\maketitle

\input{body}

\bibliographystyle{ACM-Reference-Format}
\bibliography{refs}

\input{appendix}

\end{document}

%% file: body.tex
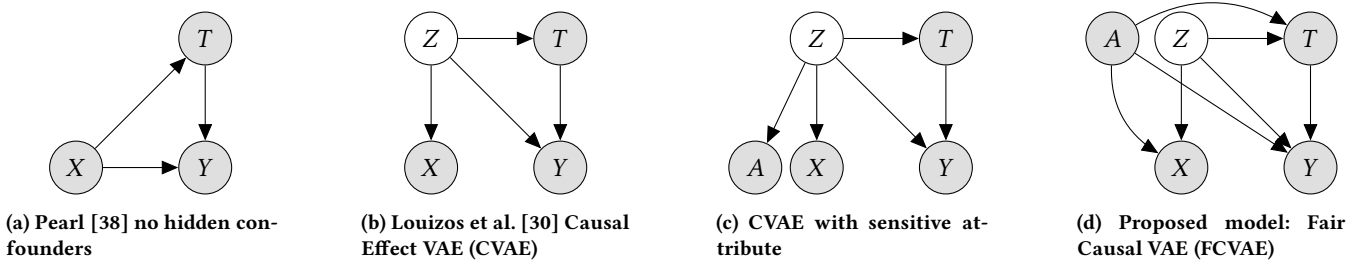
\begin{figure*}[t]
    \centering
    \begin{subfigure}[t]{0.2\textwidth}
        \centering
        \input{pearl}
	\caption{\citet{pearl2009causality} no hidden confounders}
    \label{subfig:pearl}
    \end{subfigure}%
    \hfill 
    \begin{subfigure}[t]{0.2\textwidth}
        \centering
        \input{cevae}
	\caption{\citet{louizos2017causal} Causal Effect VAE (CVAE)}
    \label{subfig:cevae}
    \end{subfigure}%
    \hfill
    \begin{subfigure}[t]{0.2\textwidth}
        \centering
        \input{cevae-sens-attr}
        \caption{CVAE with sensitive attribute}
        \label{subfig:cevae-sens}
    \end{subfigure}
	\hfill
    \begin{subfigure}[t]{0.2\textwidth}
        \centering
        \input{fair-cevae}
        \caption{Proposed model: Fair Causal VAE (FCVAE)}
        \label{subfig:cevae-fair}
    \label{fig:proposed}
    \end{subfigure}
    \caption{Various approaches causally modeling data features $X$, treatment $T$, and outcome $Y$. \ref{subfig:pearl} assumes no hidden confounders; \ref{subfig:cevae} models hidden confounders via a latent variable $Z$ to be inferred by an inference neural network (not pictured); \ref{subfig:cevae-sens} naively extends \ref{subfig:cevae} to include a sensitive attribute $A$ as an additional observation, not as a confounder. \ref{subfig:cevae-fair}, the Fair Causal VAE (ours), explicitly models the sensitive attribute as confounding the treatment $T$ and the label $Y$ in historical data.}
    \label{fig:cevae}
    \end{figure*}

\section{Introduction} \label{sec:intro}
In this work, we consider the problem of fair decision-making from biased datasets.
Much work has been done recently on the problem of fair classification \citep{zafar2015fairness,hardt2016equality,bechavod2017penalizing,agarwal2018reductions}, yielding an abundant supply of definitions, models, and algorithms for the purposes of learning classifiers whose outputs satisfy distributional constraints.
Some of the canonical problems for which these algorithms have been proposed are loan assignment \citep{hardt2016equality}, criminal risk assessment \citep{chouldechova2017fair}, and school admissions \citep{friedler2016possibility}.
However, none of these problems are fully specified by the classification paradigm.
Rather, they are decision-making problems: each problem requires an action (or ``treatment'') to be taken in the world, which in turn yields an outcome.
In other words, the central question is how to intervene in an ongoing and evolving process, rather than predict outcomes alone \citep{barabas2017interventions}.

Decision-making, i.e. learning to intervene, requires a fundamentally different approach from learning to classify: historical training data are the product of past interventions and thus provide an incomplete view of all possible outcomes. 
Only actions which were previously chosen yield observable outcomes in the training data, while the implicit counterfactual outcomes (the outcome that would have occurred had another action been taken) are never observed.
The incompleteness of this data can have great impact on learning and inference \cite{rubin1976inference}. 

It has been widely argued that biased data yields unfair machine learning systems \citep{kallus2018residual,hashimoto2018fairness,pmlr-v81-ensign18a}.
In this work we examine dataset bias through the lens of causal inference.
To understand how past decisions may bias a dataset, we first must understand how sensitive attributes may have affected the generative process which created the dataset, including the (historical) decision makers' actions (treatments) and results (outcomes).
Causal inference is well suited to this task: since we are interested in decision-making rather than classification, we should be interested in the causal effects of actions rather than correlations.
Causal inference has the added benefit of answering counterfactual queries: What would this outcome have been under another treatment? 
How would the outcome change if the sensitive attribute were changed, all else being equal? 
These questions are core to the mission of learning fair systems which aim to inform decision-making \citep{kusner2017counterfactual}.

While there is much that causal inference can offer to the field of fair machine learning, it also poses several significant challenges.
For example, the presence of \emph{hidden confounders}---unobserved factors that effect both the historical choice of treatment and the outcome---often prohibits the exact inference of causal effects.
Additionally, understanding effects at the individual level can be especially complex, particularly if the outcome is non-linear in the data and treatments.
These technical difficulties are often amplified by the problem scope of modern machine learning, where large and high-dimensional datasets are commonplace.

To address these challenges, we propose a model for fairly estimating individual-level causal effects from biased data, which combines causal modeling \citep{pearl2009causality} with approximate inference in deep latent variable models \citep{kucukelbir2017automatic,louizos2017causal}. 
Our focus on individual-level causal effects and counterfactuals provides a natural fit for application areas requiring fair policies and treatments for individuals, such as finance, medicine, and law.
Specifically, we incorporate the sensitive attribute into our model as a confounding factor, which can possibly influence both the treatment and the outcome.
This is a first step towards achieving ``fairness through awareness'' \citep{dwork2012fairness} in the interventional setting.

Our model also leverages recent advances in deep latent-variable modeling to model potential hidden confounders as well as complex, non-linear functions between variables, which greatly increases the class of relationships which it can represent.
Through experimental analysis, we show that our model can outperform non-causal models, as well as causal models which do not consider the sensitive attribute as a confounder. 
We further explore the performance of this model, showing that fair-aware causal modeling can lead to more accurate, fairer policies in decision-making systems.

\section{Background} \label{sec:causal-background}
\subsection{Causal Inference}\label{sec:causal-inference}  
We employ Structural Causal Models (SCMs), which provide a general theory for modeling causal relationships between variables \citep{pearl2009causality}.
An SCM is defined by a directed graph, containing vertices and edges, which respectively represent variables in the world and their pairwise causal relationships.
There are two types of vertices: exogenous variables $\mathcal{U}$ and endogenous variables $\mathcal{V}$.
Exogenous variables are unspecified by the model; we model them as unexplained noise distributions, and they have no parents.
Endogenous variables are the objects we wish to understand; they are descendants of endogenous variables.
The value of each endogenous variable is fully determined by its ancestors.
Each $V \in \mathcal{V}$ has some function $f_V$ which maps the values of its immediate parents to its own.
This function $f_V$ is deterministic; any randomness in an SCM is due to its exogenous variables.

In this paper, we are primarily concerned with three endogenous variables in particular: $X$, the observable \emph{features} (or covariates) of some example; $T$, a \textit{treatment} which is applied to an example; and $Y$, the \textit{outcome} of a treatment.
Our decision problem is: given an example with particular values for its features, $X=x$, what value should we assign to treatment $T$ in order to produce the best outcome $Y$?
This is fundamentally different from a classification problem, since typically we observe the result of only one treatment per example
\footnote{
{\color{black} 
Note that we use the terms \textit{treatment} and \textit{outcome} as general descriptors of a decision made/action taken and its result, respectively.
These terms are associated with an alternative theory of causal inference \citep{rubin2005causal} which can also be used to describe the methods we propose, but which we will not discuss in this paper.
}
}
.

To answer this decision problem, we need to understand the value $Y$ will take if we \textit{intervene} on $T$ and set it to value $t$.
Our first instinct may be to estimate $P(Y | T = t, X = x)$.
However, this is unsatisfactory in general.
If we are estimating these probabilities from observational data, then the fact that $x$ received treatment $t$ \textit{in the past} may have some correlation with the historical outcome $Y$.
This ``confounding'' effect---the fact that $X$  has an effect on both $T$ and $Y$ is depicted in Figure \ref{subfig:pearl}, by the arrows pointing out of $X$ into $T$ and $Y$.
For instance, in an observational medical trial, it is possible that young people are more likely to choose a treatment, and also that young people are more likely to recover.
A supervised learning model, given this data, may then overestimate the average effectiveness of the treatment on a test population. 
Broadly, to understand the effect of assigning treatment $t$, supervised learning is not enough; we need to model the functions $\{f_V\}$ of the SCM.

Once we have a fully defined SCM, we can use the $do$ operation \cite{pearl2009causality} to simulate the distribution over $Y$ given that we assign some treatment $t$---we denote this as $P(Y | do(T = t), X = x)$.
We do the $do$ through graph surgery: we assign the value $t$ to $T$ by removing all arrows going into $T$ from the SCM and setting the corresponding structural equation output to the desired value regardless of its input $f_T(\cdot) = t$. We then set $X = x$ and continue with inference of $Y$ as we normally would.

A common assumption in causal modelling is the ``no hidden confounders'' assumption, which states that there are no unobserved variables affecting both the treatment and outcome. 
We follow \citet{louizos2017causal}, and use variational inference to model confounders that are not directly observed but can be abstracted from proxies.
{\color{black} In Sec. \ref{sec:models} we consider the implications of this approach and discuss alternative assumptions.
}

\subsection{Approximate Inference}
Individual and population-level causal effects can be estimated via the \emph{do} operation when the values of all confounding variables are observed \citep{pearl2009causality}, which motivates the common no-hidden-confounders assumption in causal inference. 
However this assumption is rather strong and precludes classical causal inference in many situations relevant to fair machine learning, e.g., where ill-quantified and hard-to-observe factors such socio-economic status (SES) may significantly confound the observable data.
Therefore we follow \citet{louizos2017causal} in modeling unobserved confounders using a high dimensional latent variable $Z$ to be inferred for each observation $(X, T, Y)$.
They prove that if the full joint distribution is successfully recovered, individual treatment effects are identifiable, even in the presence of hidden confounders.
In other words, causal effects are identifiable insofar as exact inference can be carried out, and the observed covariates are sufficiently informative.

Because exact inference of $Z$ is intractable for many interesting models, we approximately infer $Z$ by variational inference, specifying $q(Z|X, T, Y)$ using a parametric family of distributions and learning the parameters that best approximate the true posterior $p(Z|X, T, Y)$ by maximizing the evidence lower bound (ELBO) of the marginal data likelihood \citep{wainwright2008graphical}.
In particular, we amortize inference by training a neural network (whose functional form is specified separately from the causal model) to predict the parameters of $q$ given $(X, T, Y)$ \cite{kingma2013auto}.
Amortized inference is much faster but less optimal than local inference \cite{kim2018semi}; alternate inference strategies could be explored for applications where the importance of accuracy in individual estimation justifies the additional computational cost.

\subsection{TARNets} \label{sec:tarnets}
TARNets \citep{shalit2016estimating} are 
{a \color{black} class of neural network}
architectures for estimating outcomes of a binary treatment. 
{\color{black}
The network comprises two separate arms---each predicts the outcomes associated with a separate treatment---that share parameters in the lower layers.
The entire network is trained end to end using gradient-based optimization, but with
}
only one arm (the one with the treatment which was actually given) receiving error signal for any given example.
The TARNet prediction of result $R$ and input variables $V$ and potential intervention $I$ is expressed by combining the shared representation function $\Phi$ with the two functions $h_0, h_1$ corresponding to the separate prediction arms.
This yields two composed functions,
\begin{equation}
\begin{aligned}
g^{I=0}_R(V, I) &= h_0(\Phi(V))\\
g^{I=1}_R(V, I) &= h_1(\Phi(V))
\end{aligned}
\end{equation}
with $h_0, h_1, \Phi$ realized as neural networks.
\citet{shalit2016estimating} explore a group-wise MMD penalty on the outputs of $\Phi$; we do not use this.

\section{Fair Causal Inference} \label{sec:problem-setup}

As stated in Sec. \ref{sec:causal-inference}, we are interested in modeling the causal effects of treatments on outcomes. 
However, when attempting to learn fairly from a biased dataset, this problem takes on an extra dimension. 
In this context, we become concerned with understanding causal effects in the presence of a \emph{sensitive attribute} (or protected attribute).
Examples include race, gender, age, or SES.
When learning from a historical data, we may believe that one of these attributes affected the observable treatments and outcomes, resulting in a biased dataset.

\citet{lum2016predict} give an example in the domain of predictive policing of how a dataset of drug crimes may become biased with respect to race through unfair policing practices.
They note that it is impossible to collect a dataset of all drug crimes in some area; rather, these datasets are really tracking drug \emph{arrests}.
Due to a higher level of police presence in heavily Black than heavily White communities, recorded drug arrests will by nature over-represent Black communities.
Therefore, a predictive policing algorithm which attempts to fit this data will continue the pattern of over-policing Black communities.
\citet{lum2016predict} provide experimental validation of this hypothesis through simulation, contrasting the output of a common predictive policing algorithm with independent, demographic-based estimates of drug use by neighborhood.
Their work shows that wrongly specifying a learning problem as one of supervised classification can lead to replicating past biases.
In order to account for this in the learning process, we should be aware of the biases which shaped the data --- which may include sensitive attributes that historically affected the treatment and/or outcome.

Using the above example for concreteness, we specify the variables at play.
The decision-making problem is: should police be sent to neighborhood $X$ at a given time?
The variables are:
\begin{itemize}
\item $A \in \{0, 1\}$: a sensitive attribute. For example the majority race of a neighborhood.
\item $T \in \{0, 1\}$: a treatment. For example the presence or absence of police in a certain neighborhood on a particular day.
\item $Y \in \mathds{R}$: an outcome. For example the number of arrests recorded in a given neighborhood on a particular day.
\item $X \in \mathds{R}^D$: $D$-dimensional observed features. For example statistics about the neighborhood, which may change day-to-day
\end{itemize}
We will represent sensitive attributes and treatments as binary throughout this paper; we recognize this is not always an optimal modeling choice in practice.
Note that the choice of treatment will causally alter the outcome---an arrest cannot occur if there are no police in the area.
Furthermore, the sensitive attribute can causally effect the outcome as well; research has shown that policing can disparately effect various races, even controlling for police presence \cite{gelman2007analysis} (the treatment in this case).

We note that in various domains, there may be more variables of interest than the ones we list here, and more appropriate causal models than those shown in Fig. \ref{fig:cevae}. 
However, we believe that the setup we describe is widely applicable
and contains the minimal set of variables to be useful for fairness-aware causal analysis.
We are interested in calculating causal effects between the above variables. In particular, we seek answers to the following three questions:
\paragraph{What is the effect of the treatment on the outcome?}
This will help us to understand which $T$ is likely to produce a favorable outcome for a given $X$.
Let us denote~$y_{T=t}(x, a) = \mathds{E}[y | do(T = t), X = x, A = a]$~as the expected conditional outcome under $T = t$, that is, the ground truth value taken by $Y$ when the treatment $T$ is assigned the value $t$, and conditioning on the values $x, a$ for the features and sensitive attribute respectively.
Then, we can express the individual effect of $T$ on $Y$ ($IE_{T \rightarrow Y}$) as 
\begin{equation}
IE_{T \rightarrow Y}(x, a) = y_{T=1}(x, a) - y_{T=0}(x, a).
\end{equation}

\paragraph{What is the effect of the sensitive attribute on the treatment?}
This allows us to understand how the treatment assignment was biased in the data.
Similarly, we can define $t_{A=a}(x) = \mathds{E}[t | do(A = a), X = x]$, which is the expected conditional treatment in the historical data when the value $a$ is assigned to the sensitive attribute. 
Then, the individual effect of $A$ on $T$ can be expressed as
\begin{equation}
IE_{A \rightarrow T}(x) = t_{A=1}(x) - t_{A=0}(x).
\end{equation}

\paragraph{What is the effect of the sensitive attribute on the outcome?}
This allows us to understand what bias is introduced into the historically observed outcome.
We can also define $y_{A=a}(x) = \mathds{E}[y | do(A = a), X = x, T=t_{A=a}(x)]$ as the expected conditional outcome under $A = a$; the ground truth value of $Y$ conditioned on the features being $x$ if the sensitive attribute were assigned the value $a$, and the treatment $T$ were assigned the ground truth value $t_{A=a}(x)$. 
Then, we can express the individual effect of $A$ on $Y$ as 
\begin{equation}
IE_{A \rightarrow Y}(x) = y_{A=1}(x) - y_{A=0}(x).
\end{equation}

{\color{black}
\subsection{Intervening on Sensitive Attributes}
There has been some disagreement around the notion of intervening on an immutable (or effectively immutable) sensitive attribute.
\citet{holland1986statistics} argue that there is ``no causation without manipulation'' --- i.e. an attribute can never be a cause; only an experience undergone can be.
Briefly stated, they argue that if the factual and counterfactual versions cannot be ``defined in principle, it is impossible to define the causal effect''.
In a counterargument, \citet{marini1988causality} claim that a ``synthesis of intrinsic and extrinsic determination [provides] a more adequate picture of causal relations'' --- meaning that both externally imposed experiences (extrinsic) and internally defined attributes (intrinsic) are valid conceptual components of a theory of causation.
We agree with this view --- that the notion of a causal effect of an immutable attribute is valid, and believe that it is particularly useful in a fairness context.

Specifically pertaining to race, some argue it is possible to understand the causal effect of an immutable attribute in terms of the effects of more manipulable attributes (proxies).
\citet{vanderweele2014causal} argue that, rather than interpreting a causal effect estimate of $A$ as a hypothetical randomized intervention on $A$, one can interpret it as a particular type of intervention on some other set of manipulable variables related to $A$ (under certain graphical and distributional assumptions on those variables). 
\citet{sen2016race} take a constructivist approach, and consider race to be composed of constituent parts, some of which \textit{can} be theoretically manipulated.
They describe several experimental designs which could estimate the effects of immutable attributes.

Another issue with intervening on sensitive attributes is that, since many are ``assigned at conception'', all observed covariates $X$ are post-treatment \citep{sen2016race} (as reflected in the design of our SCM in Fig. \ref{subfig:cevae-fair}).
In statistical analysis, a frequent approach is to ignore all post-treatment variables to avoid introducing collider biases \citep{gelman2007analysis,king1994designing}.
However, in our model, the purpose of the covariates is to deduce the true (unobserved) values of the latent $Z$ for that individual.
Therefore, when conditioning on the observed covariates, correlation of $A$ and $Z$ is the objective, rather than an undesired side effect.
This is the first step (``Abduction'') of computing counterfactuals (according to \citet{pearl2009causality}); we can think of this as adjusting for bias (of the sensitive attribute) in the $X$-generating process.
}

\section{Proposed Method} \label{sec:models}
In this section we first conceptualize and describe our proposed causal model---depicted in Fig. \ref{fig:proposed}---then discuss the parameterization of the corresponding SCMs and learning procedure.
{\color{black}
A common causal modelling approach is to define a new SCM for each problem \citet{pearl2009causality}, taking advantage of domain specific knowledge for that particular problem.
This stands in contrast to a classic machine learning (ML) approach, which aims to process data and draw conclusions as generally as possible, by automatically discovering patterns of correlation in the data.
While the causal modelling approach is capable of detecting effects the ML approach cannot, the ML approach is attractive since it provides modularity, generality and a more automated data processing pipeline.
In this work, we aim to interpolate between the two approaches by considering a single, general causal model for observational data.
Our model contains what we argue are a minimal set of fairly general causal variables for discovering treatment effects and biases in the data-generation process, allowing us to interface causally with arbitrary data that fits the proposed structure.
}

Two features of our causal model are noteworthy.
First is the explicit consideration of the sensitive attribute---a potential source of dataset bias---as a confounder, which causally affects both the treatment $T$ and the outcome $Y$.
This contrasts with approaches from outside the fairness literature (e.g. \citep{louizos2017causal}, Fig. \ref{subfig:cevae}), which in a fairness setting (Fig. \ref{subfig:cevae-sens}) would treat potential sensitive attributes as equivalent to other observed features. 
Our model accounts for the possibility that a sensitive attribute may have causal influence on the observed features, treatments and outcomes and the historical process which generated them.
It makes the sensitive attribute distinct from the other attributes of $X$, which we understand not as confounders but observed proxies.
We can think of this as a causal modeling analogue of ``fairness through awareness''.
By actively adjusting for causal confounding effects of sensitive attributes, we can build a model which accounts for the interplay between the treatment and outcome for both values of the sensitive attribute.

The other noteworthy aspect of our model is the latent variable $Z$. 
Together, $Z$ and $A$ make up all the confounding variables. 
{\color{black}
We note two important points about these confounders.
Firstly, we clarify that the model class we propose (a latent Gaussian and a deep neural network), is not necessarily the definitive model of the confounders of $T$ and $Y$; however, it is a flexible one, with numerous applications in machine learning \citep{rezende2014stochastic}.
Secondly, we note that causal inference and machine learning have different conventions around unobserved (i.e. latent) variables --- in causal inference, these variables are generally considered to be nameable objects in the world (e.g. SES, historical predjudice), whereas in machine learning they represent some unspecified (and perhaps abstract) structure in the data.
Our $Z$ follows the machine learning convention.
}

As in \citet{louizos2017causal}, $Z$  represents all the unobserved confounding variables which effect the outcomes or treatments (other than $A$).
The features $X$ can be seen as proxies (noisy observations) for the confounders ($Z, A$).
Altogether, the endogenous variables in our model are $X$, $A$, $Z$, $T$, and $Y$. 
We also have exogenous variables $\epsilon_X, \epsilon_A, \epsilon_Z, \epsilon_T, \epsilon_Y$ (not shown), each the immediate parent of (only) their respective endogenous variable. 
The structural equations are: 


\begin{align}\label{eq:structural-functions} \nonumber
&Z = f_Z(\epsilon_Z) 
&A = f_A(\epsilon_A) \\ \nonumber
&X = f_X(Z, A, \epsilon_X) 
&T = f_T(Z, A, \epsilon_T) \\ 
&Y = f_Y(Z, A, T, \epsilon_Y) 
&\medspace \epsilon_V \sim P_V(\epsilon_V) \ \forall V \in \{Z, A ,X, T, Y\}
\end{align}

{\color{black}
Since $Z$ does not necessarily refer to tangible objects in the world, it is reasonable that $Z \perp A$ in our model.
This does not prevent a characteristic such as SES (which may be correlated with $A$) from being a confounder ---
rather, $Z$ could represent the component of SES which is not based on $A$.
Since both confounders are inputs to all other variables in the SCM, the model can learn to represent variables which \textit{are} based on $A$, (e.g. SES) as a joint distribution of $Z$ and $A$.
}

With this SCM in hand, we can estimate various interventional outcomes, if we know the values of $f_V \ \forall \ V \in \{Z, A ,X, T, Y\}$.
For instance, we might estimate:
\begin{equation} \label{eq:scm-estimations}
\begin{aligned}
\mathds{E} \left[ Y | Z=z, A=a, do(T=1) \right] &= \mathds{E}_{\epsilon_Y \sim P_Y(\epsilon_Y)} [f_Y(z, a, 1, \epsilon_Y)]\\
\mathds{E} \left[ Y | Z=z, do(A=1), do(T=1) \right] &= \mathds{E}_{\epsilon_Y \sim P_Y(\epsilon_Y)} [f_Y(z, 1, 1, \epsilon_Y)]\\
\mathds{E} \left[ Y | Z=z, do(A=1) \right] &= \\
\mathds{E}_{\epsilon_Y \sim P_Y(\epsilon_Y)} \mathds{E}_{\epsilon_T \sim P_T(\epsilon_T)} &[f_Y(z, 1, f_T(z, 1, \epsilon_T), \epsilon_Y)]
\end{aligned}
\end{equation}
which are the expected values over outcomes of interventions on $T$, $T$ and $A$, and just $A$, respectively.

However, the problem with the calculations in Eq. \ref{eq:scm-estimations} is that $Z$ is unobserved, so we cannot simply condition on its value.
Rather, we observe some proxies $X$.
Since the structural equations go the other direction --- $X$ is a function of $Z$, not the other way around --- inferring $Z$ from a given $X$ is a non-trivial matter.

In summary, we need to learn two things: a generative model which can approximate the structural functions $f$, and an inference model which can approximate the distribution of $Z$ given $X$.
Following the lead of \citet{louizos2017causal}, we use variational inference parametrized by deep neural networks to learn the parameters of both of these models jointly.
In variational inference, we aim to learn an approximate distribution over the joint variables $P(Z, A, X, T, Y)$, by maximizing a variational lower bound on the log-probability of the observed data. 
As demonstrated in \citet{louizos2017causal}, the causal effects in the model become identifiable if we can learn this joint distribution.
We extend their proof in Appendix \ref{app:proofs} to show identifiability holds when including the sensitive attribute in the model (as in Fig. \ref{subfig:cevae-fair}).

{\color{black}
We discuss here the identifiability condition from \citet{louizos2017causal}.
Given some treatment $T$ and outcome $Y$, the classic ``no hidden confounders'' assumption asserts that the set of observed variables $O$ blocks all backdoor paths from $T$ to $Y$.
\citet{louizos2017causal} weaken this: they assume that there is a set of confounding variables $Z = O_Z \cup U_Z$ such that $Z$ blocks all backdoor paths from $T$ to $Y$, where $O_Z$ are observed and $U_Z$ are unobserved.
They claim that if we recover the full joint distribtuion of $p(Z, X, T, Y)$, then we can identify the causal effect $T \rightarrow Y$.
However, this is only possible if we have sufficiently informative proxies $X$.
While recovering the full joint distribution does not mean we have to measure every confounder, we do have to at least measure some proxy for each confounder.

This is a weaker assumption, but not fully general.
There may be confounding factors which cannot be inferred from the proxies $X$ --- in this case, our model will be unable to learn the joint distribution, and the causal effect will be unidentifiable.
In this case, we are back to square one; our causal estimates may be inaccurate. 
Determining the exact fairness implications of this remains an open problem --- it would depend on which confounders were missing, and which proxies were already collected.
A complicating factor is that testing for unconfoundedness is difficult, and usually requires making further assumptions \citep{tran2016model}.
Therefore we might unintentionally make unfair inferences if we are unaware that we cannot infer all confounders.
If we think this is the case, one solution is to collect more proxies.
This provides an alternative motivation for the idea of increasing fairness by measuring additional variables \citep{chen2018my}.
}

To learn a generative model of the data which is faithful to the structural model defined in Eq. \ref{eq:structural-functions}, we define distributions $p$ which will approximate various conditional probabilities in our model. We model the joint probability assuming the following factorization:
\begin{equation} \label{eq:joint-probability}
\begin{aligned}
P(Z, A, X, T, Y) = p(Z) p(A) p(X | Z, A) p(T | Z, A) p(Y | Z, A, T)
\end{aligned}
\end{equation}
Each of these $p$ corresponds to an $f$ in Eq. \ref{eq:structural-functions} --- formally, $p(V = v | W = w) = P_{\epsilon_{V}}[f_V(W, \epsilon_V)]$ for an endogenous variable $V$ and subset of endogenous variables $W$, where $\{V\}, W \subset \{Z, A, X, T, Y\}$.
For simplicity, we choose computationally tractable probability distributions for each conditional probability in Eq. \ref{eq:joint-probability}:
\begin{equation} \label{eq:distributions}
\begin{aligned}
p(Z) &= \prod_{j=1}^{D_Z} \mathcal{N}(Z_{j} | 0, 1)\\
p(A) &= Bern(A|\pi_A)\\
p(X | Z, A) &= \prod_{j=1}^{D_X} \mathcal{N}(X_{j} | \mu_X(Z, A), \sigma_X^2(Z, A) \\
p(T | Z, A) &= Bern(T | \pi_T(Z,A))
\end{aligned}
\end{equation}
where $D_Z, D_X$ are the dimensionalities of $Z$ and $X$ respectively, and $\pi_A \in [0, 1]$ is the empirical marginal probability of $A=1$ across the dataset (if this is unknown, we could use a Beta prior over that distribution; in this paper we assume $A$ is observed for every example).
For $p(Y | Z, A, T)$, we use either a Bernoulli or a Gaussian distribution, depending on if $Y$ is binary or continuous:
\begin{equation} \label{eq:distributions-Y}
\begin{aligned}
p_{binary}(Y | Z, A, T) &= Bern(Y | \pi_Y(Z,A,T))\\
p_{cont}(Y | Z, A, T) &= \mathcal{N}(Y | \mu_Y(Z,A,T), \sigma^2_Y(Z,A,T))
\end{aligned}
\end{equation}

To flexibly model the potentially complex and non-linear relationships in the true generative process, we specify several of the distribution parameters from Eqs. \ref{eq:distributions} and \ref{eq:distributions-Y} as the output of a function $g_V$, which is realized by a neural network (or TARNet \citep{shalit2016estimating}) with parameters $\theta_V$. 
We parametrize the model of $X$ with neural networks $g_X^{\mu}, g_X^{\sigma}$:

\begin{equation}\label{eq:reparamatrization-mlp}
\begin{aligned}
\mu_X(Z,A) &= g^{\mu}_X(Z, A) \\
\sigma_X^2(Z,A) &= \exp{2g^{\sigma}_X(Z, A)}
\end{aligned}
\end{equation}
 
We use TARNets \cite{shalit2016estimating} (see Sec. \ref{sec:tarnets}) to parameterize the distributions over $T$ and $Y$.
In our model, $A$ acts as the ``treatment'' for the TARNet that outputs $T$.
Likewise $A$ and $T$ are joint treatments affecting $Y$ --- our $Y$ model can be seen as a \textit{hierarchical TARNet}, with one TARNet for each value of $A$, where each TARNet has an arm for each value of $T$. 
In all, this yields the following parametrization:

\begin{equation}\label{eq:reparamatrization-tarnets}
\begin{aligned}
p_T(Z,A) &= (1 - A) \sigma(g^{A=0}_T(Z, A)) + A \sigma(g^{A=1}_T(Z, A)); \\
p_Y(Z,A,T) &= (1 - T) (1 - A) \sigma(g^{T=0,A=0}_Y(Z, A, T))\\
&\quad + T (1 - A)\sigma(g^{T=1,A=0}_Y(Z, A, T)) \\
&\quad + (1 - T) A \sigma(g^{T=0,A=1}_Y(Z, A, T)) \\
&\quad + T A \sigma(g^{T=1,A=1}_Y(Z, A, T)); 
\end{aligned}
\end{equation}
and the same for $\mu_Y(Z,A,T)$ and $\sigma_Y^2(Z,A,T)$; where $\sigma$ is the sigmoid function $\sigma(x) = \frac{1}{1 + \exp(-x)}$ and $g^{I=0}_R(V, I)$ are defined as in Sec. \ref{sec:tarnets}.

We further define an inference model $q$, to determine the values of the latent variables $Z$ given observed $X, A$. This takes the form:
\begin{equation} \label{eq:inference-model}
\begin{aligned}
q(Z | X, A) = \mathcal{N}(\mu_Z(X,A), \sigma_Z^2(X,A))
\end{aligned}
\end{equation}
where the normal distribution is reparametrized analogously to Eq. \ref{eq:reparamatrization-mlp} with networks $g_Z^{\mu}, g_Z^{\sigma}$.
Since $A$ is always observed, we do not need to infer it, even though it is a confounder.
We note that this is a different inference network from the one in \citet{louizos2017causal} --- we do not use the given treatments and outcomes in the inference model. We found it to be a simpler solution (no auxiliary networks necessary), and did not see a large change in performance. This is similar to the approach taken in \citet{parbhoo2018causal}.

To learn the parameters of this model, we can maximize the expected lower bound on the log probability of the data (the ELBO), which takes the form below, which we note is also a valid ELBO to optimize for lower-bounding the conditional log-probability of the treatments and outcomes given the data.
\begin{equation} \label{eq:elbo}
\begin{aligned}
\mathcal{L} = \sum_{i=1}^n &\mathds{E}_{q(z_i | x_i, a_i)} [\log{p(x_i | z_i, a_i)} + \log{p(t_i | z_i, a_i)} \\
&+ \log{p(y_i | z_i, a_i, t_i)} + \log{p(z_i)} - \log{q(z_i | x_i, a_i)}]
\end{aligned}
\end{equation}

\section{Related Work} \label{sec:related-work}
Our work most closely relates to the Causal Effect Variational Autoencoder \citep{louizos2017causal}.
Some follow-up work is done by \citet{parbhoo2018causal}, who suggest a purely discriminative approach using the information bottleneck.
Our model differs from this work in that they did not include a sensitive attribute in their model, and their model does not contain a ``reconstruction fidelity'' term, in this case $\log{p(x_i | z_i, a_i)}$.
Previous papers which learn causal effects using deep learning (with all confounders observed) include \citet{shalit2016estimating} and \citet{johansson2016learning}, who propose TARNets as well as some form of balancing penalty.

The intersection of fairness and causality has been explored recently. Counterfactual fairness --- the idea that a fair classifier is one which doesn't change its prediction under the counterfactual value of $X$ when $A$ is flipped --- is a major theme \citep{kusner2017counterfactual}.
Criteria for fairness in treatments are proposed in \citet{nabi2018fair}, and fair interventions are further explored in \citet{kusner2018causal}. 
\citet{zhang2018fairness} present a decomposition which provides a different way of understanding of unfairness in a causal inference model.
Other work focuses on the causal relationship between sensitive attributes and proxies in fair classification \citep{kilbertus2017avoiding}.

\citet{kallus2018residual} explore the idea of learning from biased data, making the point that a ``fair'' predictor learned on biased data may not be fair under certain forms of distributional shift, while not touching on causal ideas.
Some conceptually similar work has looked at the ``selective labels'' problem \citep{lakkaraju2018selective,dearteaga2018learning}, where only a biased selection of the data has labels available.
There has also been related work on \textit{feedback loops} in fairness, and the idea that past decisions can affect future ones, in the predictive policing \citep{lum2016predict,pmlr-v81-ensign18a} and recommender systems \citep{hashimoto2018fairness} contexts, for example.
\citet{barabas2017interventions} advocate for understanding many problems of fair prediction as ones of intervention instead.
Another variational autoencoder-based fairness model is proposed in \citet{louizos2015variational}, but with the goal of fair representation learning, rather than causal modelling.
\citet{dwork2012fairness} originated the term ``fairness through awareness'', and argued that the sensitive attribute needed to be given a place of privilege in modelling in order to reduce unfairness of outcomes.

\section{Experiments} \label{sec:experiments}

In this section we compare various methods for causal effect estimation. The three effects we are interested in are
\begin{itemize}
\item $A \rightarrow T$, the causal effect of $A$ on $T$:
\begin{equation*}
\mathds{E}(T = 1 | do(A = 1), X) - \mathds{E}(T = 1 | do(A = 0), X)
\end{equation*}
\item $A \rightarrow Y$, the causal effect of $A$ on $Y$:
\begin{equation*}
\mathds{E}(Y = 1 | do(A = 1), X) - \mathds{E}(Y = 1 | do(A = 0), X)
\end{equation*}
\item $T \rightarrow Y$, the causal effect of $T$ on $Y$: 
\begin{equation*}
\mathds{E}(Y = 1 | do(T = 1), X, A) - \mathds{E}(Y = 1 | do(T = 0), X, A)
\end{equation*}
\end{itemize}
Note that all three effects are individual-level; that is, they are conditioned on some observed $X$ (and possibly $A$), and then averaged across the dataset.

\subsection{Data}
We evaluate our model using semi-synthetic data. 
The evaluation of causal models using non-synthetic data is challenging, since a random control trial on the intervention variable is required to validate correctness --- this is doubly true in our case, where we are concerned with two different possible interventions.
Additionally, while data from random control trials for treatment variables exists (albeit uncommon), conducting a random control trial for a sensitive attribute is usually impossible.

We have adapted the IHDP dataset \citep{multisite1990enhancing,brooks-gunn_liaw_klebanov_1994}---a standard semi-synthetic causal inference benchmark---for use in the setting of causal effect estimation under a sensitive attribute.
The IHDP dataset is from a randomized experiment run by the Infant Health and Development Program (in the US), which "targeted low-birth-weight, premature infants, and provided the treatment group with both intensive high-quality child care and home visits from a trained provider" \cite{hill2011bayesian}.
Pre-treatment variables were collected from both child (e.g. birth weight, sex) and the mother at time of birth (e.g. age, marital status) and behaviors engaged in during the pregnancy (e.g. smoked cigarettes, drank alcohol), as well as the site of the intervention (where the family resided).
We choose our sensitive attribute to be mother's race, binarized as White and non-White.
We follow a similar method for generating outcomes to the Response B surface proposed in \citet{hill2011bayesian}.
However, our setup differs since we are interested in additionally modelling a sensitive attribute and hidden confounders, so there are three more steps which must be taken.
First, we need to generate outcomes $Y$ for each example for $T \in \{0, 1\}$ under the counterfactual sensitive attribute $A$. 
Second, we need to generate a treatment assignment for each example for the counterfactual value of the sensitive attribute.
Finally, we need to remove some data from the observable measurements to act as a hidden confounder, as in \citet{louizos2017causal}.

We detail our full data generation method in Appendix \ref{app:data-generation}.
We denote the outcome $Y$ under interventions $do(T=t), do(A=a$) as $y_{T=t,A=a}$.
The subroutines in Algorithms \ref{alg:ihdp-response} and \ref{alg:ihdp-treatment} generate all factual and counterfactual outcomes and treatments for each example, one for each possible setting of $A$ and/or $T$.
Values of the constants that we use for data generation can be found in Appendix \ref{app:data-generation}.

We choose our hidden confounding feature $Z$ to be birth weight.
In the second (optional) step of data generation, we choose to remove 0, 1, or 2 other features. 
Especially if we choose features which are highly correlated with the hidden confounder, this has the effect of making the estimation problem more difficult.
When removing 0 features, we do nothing.
When removing 1 feature, we remove the feature which is most highly correlated with $Z$ (head size).
When removing 2 features, we remove two features most highly correlated with $Z$ (head size \& weeks born preterm).

\subsection{Experimental Setup} \label{sec:exp-setup}

We run four different models for comparison, including the one we propose.
Since we are interested in estimating three different causal effects simultaneously ($A \rightarrow T, A \rightarrow Y, T \rightarrow Y$), we cannot compare against most standard causal inference benchmark models for treatment effect estimation.
The models we test are the following:
\begin{itemize}
\item \textbf{Counterfactual MLP (CFMLP)}: a multilayer perception (MLP) which takes the treatment and sensitive attribute as input, concatenated to $X$, and aims to predict outcome.
Counterfactual outcomes are calculated by simply flipping the relevant attributes and re-inputting the modified vector to the MLP.
A similar auxiliary network learns to predict the treatment from a vector of $X$ concatenated to $A$.
\item \textbf{Counterfactual Multiple MLP (CF4MLP)}: a set of four MLPs --- one for each combination of $(A, T) \in \{0, 1\}^2$. 
Examples are inputted into the appropriate MLP for the factual outcome, and simply inputted into another MLP for the appropriate counterfactual outcome. 
A similar pair of auxiliary networks predict treatment.
\item \textbf{Causal Effect Variational Autoencoder with Sensitive Attribute (CVAE-A, Fig. \ref{subfig:cevae-sens})}: a model similar to \citet{louizos2017causal}, but with the simpler inference model we propose.
We incorporate a sensitive attribute $A$ by concatenating $X$ to $A$ as input; counterfactuals along $A$ are taken by flipping $A$ and re-inputting the modified vector.
Counterfactuals along $T$ are taken as in \citet{louizos2017causal}.
\item \textbf{Fair Causal Effect Variational Autoencoder (FCVAE, Fig. \ref{subfig:cevae-fair})}: our proposed fair-aware causal model, with $A$ concatenated to $Z$ as confounders.
We run two versions: one where $A$ is used to help with reconstructing $X$ and inferring $Z$ from $X$ (FCVAE-1), and one where it is not (FCVAE-2).
Formally, the inference model and generative model of $X$ in FCVAE-1 are $q(Z | X, A)$ and $p(X | Z, A)$, and in FCVAE-2 are $q(Z | X)$ and $p(X | Z)$ respectively.
In both versions, $A$ is a confounder of both the treatment and the outcome.
\end{itemize}

The CFMLP is purely a classification baseline. 
It learns a mapping from input to output, estimating the conditional distribution $P(Y | X, A, T)$.
The CF4MLP shares this goal, but has a more complex architecture---it learns a disjoint set of parameters for each setting of interventions, allowing it to model completely separate generative processes.
However, it is still ultimately concerned with supervised prediction.
Furthermore, neither of these models is built to consider the impact of hidden confounders.

The CVAE-A is a model for causal inference of outcomes from treatments.
Therefore, we should expect it to perform well in estimating $T \rightarrow Y$.
It is also created to model these effects under hidden confounders.
Therefore, the difference between CVAE-A and the MLPs will tell us the improvement which comes from appropriate causal modelling rather than classification.

However, the CVAE-A does not consider the sensitive attribute $A$ as a confounder; rather, it treats it simply as another covariate of $X$.
So in comparing the FCVAE to the CVAE-A, we observe the improvement that comes from causally modelling the dataset unfairness stemming from a sensitive attribute.
In comparing the FCVAE to the MLPs, we observe the full impact of the FCVAE --- joint causal modelling of treatments, outcomes, sensitive attributes, and hidden confounders.
See Appendix \ref{app:experiments} for experimental details.

\subsection{Results} \label{sec:results}

\subsubsection{Estimating Causal Effects}

In this section, we evaluate how well the models from Sec. \ref{sec:exp-setup} can estimate the three causal effects $A \rightarrow T, A \rightarrow Y, T \rightarrow Y$.
To avoid confusion with the words \textit{treatment} and \textit{outcome}, in each of these three causal interactions, we will refer to to the causing variable as the \textit{intervention} variable, and the affected variable as the \textit{result} variable.
To evaluate how well our model can estimate causal effects, we use PEHE: Precision in Estimation of Heterogeneous Effects \cite{hill2011bayesian}.
This is calculated as: $PEHE = \sqrt[]{\mathds{E}[( (r_1 - r_0) - (\hat{r}_1 - \hat{r}_0)) ^2]}$, where $r_i$ is the ground truth value of result from the intervention $i$, and $\hat{r}_i$ is our model's estimate of that quantity.
PEHE measures our ability to model both the factual (ground truth) and the counterfactual results.

\begin{table}[]
\begin{tabular}{llll}
\hline
Model   & A $\rightarrow$ T             & T $\rightarrow$ Y             & A $\rightarrow$ Y              \\ \hline
CFMLP & 0.681 $\pm$ 0.00 & 4.51 $\pm$ 0.13 & 3.28 $\pm$ 0.07 \\
CF4MLP & 0.667 $\pm$ 0.00 & 4.58 $\pm$ 0.13 & 3.71 $\pm$ 0.09 \\
CVAE-A  & 0.665 $\pm$ 0.00 & \textbf{3.80} $\pm$ 0.10 & 3.04 $\pm$ 0.06 \\
FCVAE-1 & \textbf{0.659} $\pm$ 0.00  & \textbf{3.82} $\pm$ 0.11 & \textbf{2.88}$\pm$ 0.06  \\
FCVAE-2 & \textbf{0.659} $\pm$ 0.00 & \textbf{3.81} $\pm$ 0.11  & \textbf{2.78 }$\pm$ 0.06 \\ 
 \hline
\end{tabular}
\caption{PEHE for each model on IHDP data (no extra features removed). Mean and standard errors shown, as calculated over 500 random seedings.}
\label{table:ihdp-pehe-0}
\end{table}

\begin{table}[]
\begin{tabular}{llll}
\hline
Model   & A $\rightarrow$ T             & T $\rightarrow$ Y             & A $\rightarrow$ Y              \\ \hline
CFMLP & 0.675 $\pm$ 0.00 & 4.30 $\pm$ 0.11 & 3.42 $\pm$ 0.08 \\
CF4MLP & \textbf{0.661} $\pm$ 0.00 & 4.37 $\pm$ 0.11  & 3.89 $\pm$ 0.07\\
CVAE-A  & 0.672 $\pm$ 0.00 & \textbf{4.05}  $\pm$ 0.10 & 3.53 $\pm$ 0.07 \\
FCVAE-1 & 0.663 $\pm$ 0.00  & \textbf{4.00} $\pm$ 0.10  & \textbf{3.39} $\pm$ 0.08 \\
FCVAE-2 & 0.663 $\pm$ 0.00 & \textbf{3.99} $\pm$ 0.10  & \textbf{3.25} $\pm$ 0.07 \\ 
 \hline
\end{tabular}
\caption{PEHE for each model on IHDP data (1 most informative feature removed). Mean and standard errors shown, as calculated over 500 random seedings.}
\label{table:ihdp-pehe-1}
\end{table}

\begin{table}[]
\begin{tabular}{llll}
\hline
Model   & A $\rightarrow$ T             & T $\rightarrow$ Y             & A $\rightarrow$ Y              \\ \hline
CFMLP & 0.666 $\pm$ 0.00 & 6.03 $\pm$ 0.21 & 4.30 $\pm$ 0.12 \\
CF4MLP & \textbf{0.659} $\pm$ 0.00 & 5.77 $\pm$ 0.18 & 4.59 $\pm$ 0.10 \\
CVAE-A  & 0.672 $\pm$ 0.00 & \textbf{5.46}  $\pm$ 0.18 & 4.19 $\pm$ 0.10  \\
FCVAE-1 & \textbf{0.659} $\pm$ 0.00  & \textbf{5.40} $\pm$ 0.18  & \textbf{4.07}$\pm$ 0.11 \\
FCVAE-2 & \textbf{0.659} $\pm$ 0.00 & \textbf{5.39} $\pm$ 0.18  & \textbf{3.95} $\pm$ 0.10 \\ 
 \hline
\end{tabular}
\caption{PEHE for each model on IHDP data (2 most informative features removed). Mean and standard errors shown, as calculated over 500 random seedings. Lower is better.}
\label{table:ihdp-pehe-2}
\end{table}

In Tables \ref{table:ihdp-pehe-0}-\ref{table:ihdp-pehe-2}, we show the PEHE for each of the models described in Sec. \ref{sec:exp-setup}, for each causal effect of interest.
Each table shows results for a version of the dataset with 0-2 of the most informative features removed (as measured by correlation with the hidden confounder).
Therefore, the easiest problem is with zero features removed, the hardest is with two.
Note that in IHDP, $Y \in \mathds{R}$.

Generally, as expected, we observe that the causal models achieve lower PEHE for most estimation problems.
Also as expected, we observe that that the PEHE for the more complex estimation problems ($A \rightarrow Y, T \rightarrow Y$) increases as the most useful proxies are removed from the data.
We suspect there is less variation in the results for $A \rightarrow T$ since it is a simpler problem: there are no extra confounders (other than $Z$) or mediating factors to consider.

We find that our model (the FCVAE) compares favorably to the other models in this experiment.
We see that in general, the fair-aware models (FCVAE-1 and FCVAE-2) have lower PEHE than all other models when estimating the causal effects relating to the sensitive attribute ($A \rightarrow Y, A \rightarrow T$).
Furthermore, the FCVAE also performs similarly to the CVAE-A at $T \rightarrow Y$ estimation as well, demonstrating a slight improvement (at least in the more difficult 1, 2 features removed cases).

One interesting note is that FCVAE-1 (where $A$ is used in reconstruction of $X$ and in inference of $Z$) and FCVAE-2 seem to perform similarly, with FCVAE-2 being slightly better, if anything.
This may seem surprising at first, since one might imagine that using $A$ would allow the model to learn better representations of $X$, particularly for the purpose of doing counterfactual inference across $A$.

To explore this further, we examine in table \ref{table:ihdp-mizx} the latent representations $Z$ learned by each model in terms of their encoder mutual information between $Z$ and $X$, which is calculated as $KL(q(Z | X) || p(Z))$, the KL-divergence from the encoder posterior to the prior.
This quantity is roughly the same for both versions of the FCVAE, implying that the inference network $q(Z|\cdot)$ does not leverage the additional information provided by $A$ in its latent code $Z$. 
This is in fact sensible because FCVAE has access to $A$ as an observed confounder in modeling the structural equations.
We also noticed that CVAE contains about one bit of extra information in its latent code, implying some degree of success in capturing relevant information about $A$ in $Z$.
But if CVAE models all confounders during inference, why does it underperform relative to FCVAE estimating the downstream causal effects, especially $A \rightarrow Y$?
By making explicit the role of $A$ as confounder, we hypothesize that FCVAE can learn the interventional distributions with respect to $A$ (e.g., $p(Y|T, do(A=a), Z))$ rather than the conditional distributions of CVAE (e.g., $p(Y|T, Z(A)))$; we suspect that the gating mechanism of the TARNet implementation of the structural equations to be important in this regard.

\begin{table}[]
\begin{tabular}{ll}
\hline
Model & $KL \left[ q(z|\cdot) || p(z) \right] $  \\ \hline
CVAE-A  & 4.28 $\pm$ 0.10 \\
FCVAE-1 & 3.50 $\pm$ 0.12 \\
FCVAE-2 & 3.53 $\pm$ 0.12 \\ 
 \hline
\end{tabular}
\caption{KL divergence from the encoder posterior $q(z|\cdot)$ to prior $p(z)$ after training on IHDP; equivalent to encoder mutual information \cite{alemi2018fixing}. 
CVAE and FCVAE-1 use $(X, A)$ as input to encoder, while FCVAE-2 uses $X$ only. 
Mean and standard errors shown, as calculated over 500 random seedings.}
\label{table:ihdp-mizx}
\end{table}

\subsubsection{Learning a Treatment Policy} \label{sec:treatment-policy}

The next natural question is: how does estimating these causal effects contribute to a fair decision-making policy?
We examine two dimensions of this.
We define a \textit{policy} $\pi: X, A \rightarrow T$ as a function which maps inputs (features and sensitive attribute) to treatments.
We suppose the goal is to assign treatments $T$ using a policy $\hat T = \pi(X, A)$ that maximizes its expected \textit{value} $V(\pi)$, defined here as the expected outcome $Y$ it achieves over the data, i.e. $V(\pi) = \mathds{E}_{x, a}[Y | do(T=\pi(x, a)), A=a, X=x]$.
For example, we could imagine the treatments to be various medications, and the outcome to be some health indicator (e.g. number of months survived post-treatment).

We can derive a policy from an outcome prediction model simply by outputting the predicted argmax value over treatments,
i.e. $\pi(x, a) = \argmax_{t \in T} \mathds{E}_{\hat{Y}}[\hat{Y} | do(T=t), A=a, X=x]$,
where $\hat{Y}$ is the model's prediction of the true outcome $Y$.
The optimal policy $\pi^\star(x, a )= \argmax_{t \in T} \mathds{E}_Y[Y | do(T=t), A=a, X=x]$ takes the argmax over ground truth outcomes every time.

First, we look at the mean \textit{regret} of the policy $\pi$, which is the difference between its achieved value and the the value of the optimal policy: $R(\pi) = V(\pi^\star) - V(\pi)$.
{\color{black}
We note that in general, a policy's regret is not easy to compute or bound without assumptions on the outcome distribution in the data.
}
In Table \ref{table:ihdp-regret}, we display the expected regret values for the learned policies. 
We observe that the fair-aware model achieves lower regret than the unaware causal model, and much lower regret than the non-causal models, for both the easier and more difficult settings of the IHDP data.

\begin{table}[]
\begin{tabular}{llll}
\hline
Model   & 0 removed            & 1 removed            & 2 removed              \\ \hline
CFMLP & 0.37 $\pm$ 0.02 & 0.42 $\pm$ 0.02 & 0.81 $\pm$ 0.04 \\
CF4MLP & 0.31 $\pm$ 0.02 & 0.43 $\pm$ 0.02 & 0.59 $\pm$ 0.02 \\
CVAE-A  & 0.21 $\pm$ 0.01 & 0.38 $\pm$ 0.01 & 0.59  $\pm$ 0.02 \\
FCVAE-1 & \textbf{0.19} $\pm$ 0.01 & \textbf{0.36} $\pm$ 0.01  & \textbf{0.55} $\pm$ 0.02  \\
FCVAE-2 & \textbf{0.19} $\pm$ 0.01 & \textbf{0.35} $\pm$ 0.01 & \textbf{0.55} $\pm$ 0.02\\ 
 \hline
\end{tabular}
\caption{Regret for each model's policy on IHDP data with 0, 1, or 2 of the most useful covariates removed. Mean and standard errors shown, as calculated over 500 random seedings. Lower regret is better.}
\label{table:ihdp-regret}
\end{table}

Next, we attempt to measure the policy's fairness.
Most fairness metrics are designed for evaluating classification, not for intervention.
However, \citet{chen2018my} explore an idea which is easily adjusted to the interventional setting: that an algorithm is unfair if it is much less accurate on one subgroup.
Here, we adapt this notion to evaluate treatment policy fairness.

For any $x$, let us say the policy $\pi$ is \textit{accurate} if it chooses the treatment which in fact yields the best outcome for that individual; i.e. if $\pi(x, a) = \pi^\star(x, a)$.
We can define the \textit{accuracy} of the policy $Acc(\pi) = \mathds{E}_{x, a}[\mathds{1}(\pi(x, a) = \pi^\star(x, a))]$, where $\mathds{1}$ is an indicator function.
We can define the subgroup accuracy $Acc_{\alpha}$ as accuracy calculated while conditioning (not intervening) on a particular value $\alpha$ of $A$: $Acc_\alpha(\pi) = \mathds{E}_{x | A =\alpha}[\mathds{1}(\pi(x, \alpha) = \pi^\star(x, \alpha))]$.
We condition rather than intervene on $A$ here since we are interested in measuring the impact of the policy on real, existing populations, rather than hypothetical ones.
Finally, to evaluate the fairness of the policy, we can look at the \textit{accuracy gap}: $| Acc_1(\pi) - Acc_0(\pi) |$.
If this is high, the model is more unfair, since the policy has been more successful at modelling one group than the other, and is much more consistently choosing the correct treatment for individuals in that group.

In Table \ref{table:ihdp-acc-gap} we display the accuracy gaps for our models and baselines on the IHDP dataset.
We observe that the FCVAE achieves a smaller accuracy gap than those which do not consider the effect of the sensitive attribute.
This is an encouraging sign that by understanding the confounding influence of sensitive attributes in biasing historical datasets, we can learn treatment policies which are more accurate for all subgroups of the data.

\begin{table}[]
\begin{tabular}{llll}
\hline
Model   & 0 removed            & 1 removed            & 2 removed              \\ \hline
CFMLP & 0.042 $\pm$ 0.002 & 0.033 $\pm$ 0.002 & 0.062 $\pm$ 0.002 \\
CF4MLP & 0.034 $\pm$ 0.002 & 0.038 $\pm$ 0.002 & 0.054 $\pm$ 0.002 \\
CVAE-A  & 0.033 $\pm$ 0.001 & \textbf{0.028} $\pm$ 0.001 & 0.051  $\pm$ 0.002 \\
FCVAE-1 & \textbf{0.031} $\pm$ 0.001 & \textbf{0.028} $\pm$ 0.001  & \textbf{0.046} $\pm$ 0.001  \\
FCVAE-2 & \textbf{0.030} $\pm$ 0.001 & \textbf{0.027} $\pm$ 0.001 & \textbf{0.047} $\pm$ 0.001 \\ 
 \hline
\end{tabular}
\caption{Accuracy gap for each model's policy on IHDP data with 0, 1, or 2 of the most useful covariates removed. Mean and standard errors shown, as calculated over 500 random seedings. Lower gap is more fair.}
\label{table:ihdp-acc-gap}
\end{table}

\section{Discussion} \label{sec:conclusions}
In this paper, we proposed a causally-motivated model for learning from potentially biased data.
We emphasize the importance of modeling the potential confounders of historical datasets: we model the sensitive attribute as an observed confounder contributing to dataset bias, and leverage deep latent variable models to approximately infer other hidden confounders.

In Sec. \ref{sec:treatment-policy}, we demonstrated how to use our model to learn a simple treatment policy from data which assigns treatments more accurately and fairly than several causal and non-causal baselines.
Looking forward, the estimation of sensitive attribute causal effects suggests several compelling new research directions, which we non-exhaustively discuss here:
\begin{itemize}
\item \textbf{Counterfactual Fairness:} 
Our model learns outcomes for counterfactual values of both $T$ and $A$.
This means we could choose to implement a policy where we assess everyone under the same value $a'$, by assigning treatments to all individuals, no matter their original value $a$ of $A$, based on the inferred outcome distribution $P(Y | do(A=a'), X, T)$.
Such a policy respects the definition of \textit{counterfactual fairness} proposed by \citet{kusner2017counterfactual}, which requires invariance to counterfactuals in $A$ at the individual level.  
\item \textbf{Path-Specific Effects:}
Our model allows us to decompose $A \rightarrow Y$ into direct and indirect effects through mediation analysis of $T$ \citep{robins1992identifiability}.
By estimating this decomposition, we could learn a policy which respects \textit{path-specific} fairness, as proposed by \citet{nabi2018fair}.
\item \textbf{Analyzing Historical Bias:}
Estimating causal effects between $A$, $T$, and $Y$ permits for the analysis and comparison of bias in historical datasets.
For instance, the effect $A \rightarrow T$ is a measure of bias in a historical policy, and the effect $A \rightarrow Y$ is a measure of bias in whatever system historically generated the outcome.
This could serve as the basis of a \textit{bias auditing technique} for data scientists.
\item \textbf{Data Augmentation:}
The absence of data (especially not-at-random) has strong implications for downstream modeling in both fairness \citep{kallus2018residual} and causal inference \citep{rubin1976inference}.
Our model outputs counterfactual outcomes for both $A$ and $T$, which could be used for \textit{fair missing data imputation} \citep{van2018flexible,sterne2009multiple}.
This could in turn enable the application of simpler methods like supervised learning to interventional problems. 
\item \textbf{Fair Policies Under Constraints:}
In this paper, we consider an approach to fairness where understanding dataset bias is paramount, rather than the more common fairness-accuracy constraint-based tradeoff \citep{hardt2016equality,menon2017cost}.
However, in some domains we may be interested in policies which satisfy a \textit{fairness constraint} (e.g., the same distribution of treatments are given to each group). Estimating the underlying causal effects would be useful for constrained policy learning.
\item \textbf{Incorporating Prior Knowledge:}
Graphical models (both probabilistic and SCM) permit the specification of \textit{prior knowledge} when modeling data, and provide a framework for inference that balances these beliefs with evidence from the data.
This is a powerful fairness idea---we may believe a priori that a dataset \textit{should} look a certain way if not for some bias.
In the context of a fair machine learning pipeline that considers many datasets, this relates to the AutoML task of learning distributions over datasets that share global parameters \cite{edwards2016towards}.  
\end{itemize}

In automated decision making, the focus on intervention over classification \cite{barabas2017interventions} suggests the more equitable deployment of machine learning when only biased data are available, but also raises significant technical challenges.
We believe causal modeling to be an invaluable tool in addressing these challenges, and hope that this paper contributes to the discussion around how best to understand and make predictions from existing datasets without replicating existing biases.

%% file: pearl.tex
\begin{tikzpicture}

  \node[obs]                   (X)      {$X$} ; %
  \node[obs, right=of X]       (Y)      {$Y$} ; %
  \node[obs, right=of Y, above=of Y]       (T)      {$T$} ; %

  \edge{X}{Y, T}
  \edge{T}{Y}


\end{tikzpicture}

%% file: cevae.tex
\begin{tikzpicture}

  \node[obs]                   (X)      {$X$} ; %
  \node[latent, above=of X]    (Z)      {$Z$} ; %
  \node[obs, right=of X]       (Y)      {$Y$} ; %
  \node[obs, right=of Y, above=of Y]       (T)      {$T$} ; %

  \edge{Z}{X, Y, T}
  \edge{T}{Y}


\end{tikzpicture}

%% file: cevae-sens-attr.tex
\begin{tikzpicture}

  \node[obs]                   (X)      {$X$} ; %
  \node[obs, left=0.1 of X]        (A)      {$A$} ; %
  \node[latent, above=of X]    (Z)      {$Z$} ; %
  \node[obs, right=of X]       (Y)      {$Y$} ; %
  \node[obs, right=of Y, above=of Y]       (T)      {$T$} ; %

  \edge{Z}{X, Y, T, A}
  \edge{T}{Y}


\end{tikzpicture}

%% file: fair-cevae.tex
\begin{tikzpicture}

  \node[obs]                   (X)      {$X$} ; %
  \node[latent, above=of X]    (Z)      {$Z$} ; %
  \node[obs, left=0.2 of Z]        (A)      {$A$} ; %
  \node[obs, right=of X]       (Y)      {$Y$} ; %
  \node[obs, right=of Y, above=of Y]       (T)      {$T$} ; %

  \edge{Z}{X, Y, T}
  \edge[bend right] {A} {X}
  \edge[bend left] {A} {T}
  \edge{A}{Y}
  \edge{T}{Y}


\end{tikzpicture}

%% file: appendix.tex
\clearpage
\appendix
\section{Data Generation} \label{app:data-generation}
\begin{algorithm}[tb]\captionsetup{labelfont={sc,bf}}
    \caption{\textsc{GenerateIHDP}: Semi-synthetic Data Generation Algorithm for Fair Causal Inference}
   \label{alg:ihdp-main}
\begin{algorithmic}
	\STATE {\bfseries Step 1:} Remove all children from the dataset with non-white mothers who received the original treatment (as in \citet{hill2011bayesian}).
    \STATE {\bfseries Step 2:} Optional: Remove extra features from $X$.
   \STATE {\bfseries Step 3:} Normalize data (for each feature of $X$, subtract mean and divide by standard deviation).
   \STATE {\bfseries Step 4:} Remove some features $Z$ from the data to act as unobserved confounders.
   \STATE {\bfseries Step 5:} Remove some feature $A$ from the data to act as the sensitive attribute.
   \STATE {\bfseries Step 6:} Sample factual and counterfactual outcomes $\{y_{T=t,A=a} \ \forall t, a\} = \textsc{GenerateOutcomes(X, Z)}$.
   \STATE {\bfseries Step 7:} Sample factual and counterfactual treatments $\{t_{A=a} \ \forall a\} = \textsc{GenerateTreatments(Z)}$.
   \STATE {\bfseries Return} $Z, A, X, \{y_{T=t,A=a} \ \forall t, a\}, \{t_{A=a} \ \forall a\}$
\end{algorithmic}
\end{algorithm}

\begin{algorithm}[tb]\captionsetup{labelfont={sc,bf}}
    \caption{\textsc{GenerateOutcomes}: Generate outcomes for each value of the treatment and sensitive attribute (style of \citep{hill2011bayesian}, Resp. B)}
   \label{alg:ihdp-response}
\begin{algorithmic}
	\STATE {\bfseries Input:} Features $X$, unobserved confounders $Z$
   \STATE Let $[X, Z]$ denote the horizontal concatenation of $X$ and $Z$, and let the offset matrix $W$ be the shape of $[X, Z]$ with 0.5 in every position.
   	\STATE Sample $\beta \sim P_{\beta}$, choose $\omega, \beta_A \in \mathds{R}$.
   \STATE  Sample $y_{T=0,A=0} \sim \mathcal{N}(\exp(([X, Z] + W)\beta^T), 1)$
  \STATE  Sample $y_{T=1,A=0} \sim \mathcal{N}([X, Z]\beta^T - \omega, 1)$
  \STATE Sample $y_{T=0,A=1} \sim \mathcal{N}(\exp(([X, Z] + W)\beta^T) + \beta_A, 1)$
  \STATE Sample $y_{T=1,A=1} \sim \mathcal{N}([X, Z]\beta^T - \omega + \beta_A), 1)$
  \STATE \textbf{Return} \{$y_{T=0,A=0}, y_{T=1,A=0}, y_{T=0,A=1}, y_{T=1,A=1}$\}
\end{algorithmic}
\end{algorithm}

\begin{algorithm}[tb]\captionsetup{labelfont={sc,bf}}
    \caption{\textsc{GenerateTreatments}: Generate treatments for each value of the sensitive attribute}
   \label{alg:ihdp-treatment}
\begin{algorithmic}
	\STATE {\bfseries Input:} Unobserved confounders $Z$
	\STATE  Choose $\alpha_0, \alpha_1 \in [0, 1], \zeta \in \mathds{R}$.
    \STATE Let $p_{A=0} = Clip(\alpha_0 + \zeta Z, 0, 1), p_{A=1} = Clip(\alpha_1 + \zeta Z, 0, 1)$
   \STATE Sample $t_{A=0} \sim Bern(p_{A=0})$.
   \STATE Sample $t_{A=1} \sim Bern(p_{A=1})$.
   \STATE \textbf{Return} $\{t_{A=0}, t_{A=1}\}$
\end{algorithmic}
\end{algorithm}

We detail our dataset generation progess in Algorithm \ref{alg:ihdp-main}.
We denote the outcome $Y$ under interventions $do(T=t), do(A=a$) as $y_{T=t,A=a}$.
The subroutines in Algorithms \ref{alg:ihdp-response} and \ref{alg:ihdp-treatment} generate all factual and counterfactual outcomes and treatments for each example, one for each possible setting of $A$ and/or $T$.
In Algorithm \ref{alg:ihdp-main}, we have several undefined constant variables. We use the following values for those variables:
\begin{itemize}
\item $\beta \sim P_{\beta}$:
\begin{itemize}  
\item for continuous variables, \\
\ $\beta_i \sim Cat([0,.1,.2,.3,.4],[.5, .125, .125, .125, .125])$
\item for binary variables $\beta_i \sim Cat([0,.1,.2,.3,.4], [.6, .1, .1, .1, .1])$
\item for $Z$, $\beta_i = Cat([.4,.6], [.5,.5])$
\end{itemize}
where $Cat(x, p)$ selects values from $x$ according to the array of probabilities $p$.
\item $\beta_A = 6$
\item $\omega = -11$
\item $\alpha_0, \alpha_1, \zeta = 0.7, 0.4, 0.1$
\end{itemize}
We also use the function $Clip$, which is defined as:
\begin{equation}
Clip(x, m, M) = \min(\max(x, m), M); x, m, M \in \mathds{R}
\end{equation}

\section{Identifiability of Causal Effects} \label{app:proofs}

Here we show that if we can successfully recover the joint distribution $P(Z, A, X, T, Y)$, we can recover all three treatment effects we are interested in:
\begin{enumerate}
\item The effect of $T$ on $Y$ ($T \rightarrow Y$): $\mathds{E}(Y | do(T = 1), X, A) - \mathds{E}(Y | do(T = 0), X, A)$
\item The effect of $A$ on $T$ ($A \rightarrow T$): $\mathds{E}(T | do(A = 1), X) - \mathds{E}(T | do(A = 0), X)$
\item The effect of $A$ on $Y$ ($A \rightarrow Y$): $\mathds{E}(Y | do(A = 1), X) - \mathds{E}(Y | do(A = 0), X)$
\end{enumerate}
Our proof will closely follow \citet{louizos2017causal}.
For each effects, it will suffice to show that we can recover the first term on the right-hand side of each expression. (The argument for the second term is the same).
We will show only the proof for the effect of $T$ on $Y$ --- the others are very similar. \\

{\bf Theorem.} Given the causal model in Fig. \ref{subfig:cevae-fair}, if we recover the joint distribution $P(Z, A, X, T, Y)$, then we can recover $\mathds{E}(Y | do(T = 1), X, A)$.

{\it Proof.} We have that
\begin{equation}
\begin{aligned}
P(Y | do(T = 1), X, A) &= \int_Z P(Y | do(T = 1), X, A, Z) P(Z | do(T = 1), X, A)\\
\end{aligned}
\end{equation}
By the $do$-calculus, we can reduce further:
\begin{equation}
\begin{aligned}
&= \int_Z P(Y | T = 1, X, A, Z) P(Z | X, A)
\end{aligned}
\end{equation}
If we know the joint distribution $P(Z, A, X, T, Y)$, we can identify the value of each term in this expression; hence we can identify the value of the whole expression. 
\hfill \qedsymbol\\

\section{Experimental details} \label{app:experiments}
We run each model on 500 distinct data seed/model seed pairs, in order to get robust confidence estimates on the error of each model.
We parametrize each function in our causal model with a neural network. Our networks between $X$ and $Z$ have a single hidden layer of 20 hidden units. 
The size of the learned hidden confounder $Z$ was 10 units. 
Each of our TARNets consist of a network outputting a shared representation, and two networks making predictions from that representation. 
Each of these network have 1 hidden layer with 100 hidden units.
The size of the shared representation in the TARNets was 20 units.
For simplicity, we set $g_X^{\sigma} = 1$ for all experiments (but not $g_Z^{\sigma}$)---this amounts to assuming unit variance for the data $X$, a sensible assumption because they are normalized during pre-processing.
We used ELU non-linear activations \cite{clevert2015fast}.
We trained our model with ADAM \citep{kingma2014adam} with a learning rate of 0.001, calculating the ELBO on a validation set and stopping training after 10 consecutive epochs without improvement.
We sample 10 times from the posterior $q(Z|\cdot)$ at both training and test time for each input example.
At training time we compute the average ELBO across the ten samples, while at test time we use the average prediction.